 \documentclass[pmlr,twocolumn,10pt]{jmlr} 

\usepackage{graphicx}

\usepackage{booktabs}
\usepackage{hyperref}
\usepackage{textcomp}
\usepackage{siunitx}
\usepackage{chngcntr}

\usepackage[switch]{lineno}

\theorembodyfont{\upshape}
\theoremheaderfont{\scshape}
\theorempostheader{:}
\theoremsep{\newline}

\jmlrvolume{259}
\jmlryear{2024}
\jmlrsubmitted{LEAVE UNSET}
\jmlrpublished{LEAVE UNSET}
\jmlrworkshop{Machine Learning for Health (ML4H) 2024} 

\title[Federated Class-Heterogeneous Learning for Chest X-Ray Classification]{From Isolation to Collaboration: Federated Class-Heterogeneous Learning for Chest X-Ray Classification}

\author{%
\Name{Pranav Kulkarni}$^{1,2}$\Email{pkulkarni@som.umaryland.edu} \\
\Name{Adway Kanhere}$^{1,2}$\Email{akanhere@som.umaryland.edu} \\
\Name{Paul H. Yi}$^{3}$\Email{paul.yi@stjude.org} \\
\Name{Vishwa S. Parekh}$^{1}$\Email{vparekh@som.umaryland.edu}\\
\addr $^{1}$University of Maryland School of Medicine, Baltimore, MD \\
\addr $^{2}$University of Maryland Institute for Health Computing, North Bethesda, MD \\
\addr $^{3}$St. Jude Children's Research Hospital, Memphis, TN
}

\begin{document}

\maketitle

\begin{abstract}
Federated learning (FL) is a promising paradigm to collaboratively train a global chest x-ray (CXR) classification model using distributed datasets while preserving patient privacy. A significant, yet relatively underexplored, challenge in FL is class-heterogeneity, where clients have different sets of classes. We propose surgical aggregation, a FL method that uses selective aggregation to collaboratively train a global model using distributed, class-heterogeneous datasets. Unlike other methods, our method does not rely on the assumption that clients share the same classes as other clients, know the classes of other clients, or have access to a fully annotated dataset. We evaluate surgical aggregation using class-heterogeneous CXR datasets across IID and non-IID settings. Our results show that our method outperforms current methods and has better generalizability.
\end{abstract}


\paragraph*{Data and Code Availability}

All code and data used in this study is publicly available. The NIH Chest X-ray 14 dataset (\url{https://nihcc.app.box.com/v/ChestXray-NIHCC}) is open access. The CheXpert (\url{https://stanfordmlgroup.github.io/competitions/chexpert/}) and MIMIC-CXR-JPG (\url{https://physionet.org/content/mimic-cxr-jpg/2.1.0/}) datasets are credentialed access. Our code is available at \url{https://github.com/BioIntelligence-Lab/SurgicalAggregation}.

\paragraph*{Institutional Review Board (IRB)}

This retrospective study used publicly available datasets and was acknowledged by our IRB as exempt.

\section{Introduction}

Federated learning (FL) has emerged as a promising way to collaboratively train deep learning (DL) models for chest x-ray (CXR) classification using private datasets distributed across different institutions \citep{chowdhury2022review,sheller2020federated}. Consisting of a central server and clients, FL can collaboratively train a global model by aggregating the weights of local models communicated by each client using a strategy (e.g., FedAvg) while preserving patient privacy \citep{mcmahan2017communication}.

A fundamental challenge in FL is heterogeneity, where differences in data distribution among clients in a network impacts model optimization \citep{li2020federated}. This is relevant in medical imaging because each institution uses different acquisition parameters and image processing techniques, resulting in CXR datasets that are heterogeneous and not independently and identically distributed (IID). Moreover, each institution specializes in different tasks due to the costly and time-consuming process of manual annotation, resulting in partially annotated, \emph{class-heterogeneous} datasets \citep{miao2023fedseg}. In other words, annotations present in one dataset may not be present in another. For example, the NIH Chest X-Ray 14 and CheXpert datasets contain 14 and 13 disease labels respectively, with only seven in common \citep{wang2017chestx,irvin2019chexpert}. 

Despite its prevalence in medical imaging, class-heterogeneity is a relatively underexplored topic in FL literature. While several groups have proposed techniques to tackle heterogeneity, they assume that all clients share the same set of classes \citep{li2021fedbn,li2019fedmd,karimireddy2020scaffold}. However, in the real world, this may not always be the case. While some groups have attempted to solve class-heterogeneity, their techniques assume that clients know the classes of other clients \citep{miao2023fedseg,dong2022federated} or use knowledge distillation with a public dataset in the central server that is either fully annotated \citep{gudur2021resource} or unlabeled \citep{gong2021ensemble,gong2022federated}. Others avoid this challenge using personalized FL (PFL) techniques with client-specific classification heads \citep{kanhere2024privacy,kulkarni2022competition}.

\begin{figure}
  \centering
  \includegraphics[width = \linewidth]{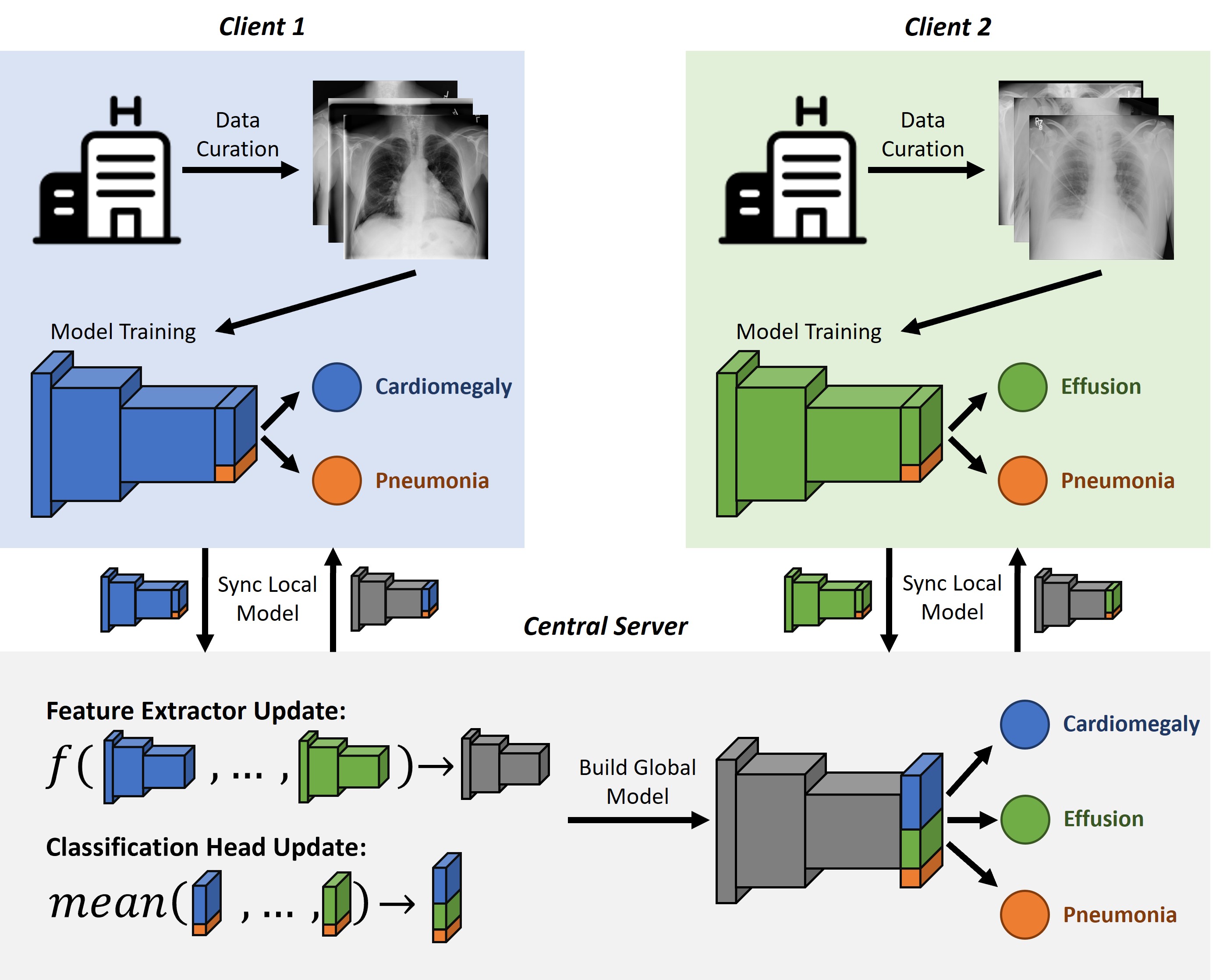}
  \caption{An overview of surgical aggregation. Suppose two sites want to collaboratively train a CXR classification model using FL. The first site works on cardiomegaly and pneumonia, while the second works on effusion and pneumonia. Using surgical aggregation, the central server can learn a global model to predict all three labels.}
  \label{fig:framework_overview}
\end{figure}

In this work, we propose \emph{surgical aggregation} as a FL method to collaboratively train a global CXR classification model using distributed, class-heterogeneous datasets (\figureref{fig:framework_overview}). Our method uses selective aggregation to dynamically build a global classification head that can predict the presence of all classes. Unlike other FL methods, surgical aggregation does not rely on the assumption that all clients must share the same classes, know the classes of other clients, nor have access to a fully annotated public dataset. It enables a client to either share, partially share, or not share any labels with other clients. The purpose of this study is to evaluate surgical aggregation for training a global model using distributed, class-heterogeneous CXR datasets across IID and non-IID settings.

\section{Methods} \label{sec:methods}

\subsection{Surgical Aggregation}

Surgical aggregation is based on the selective aggregation of weights communicated by clients to dynamically build a global classification head that predicts all classes available in the network from distributed, class-heterogeneous datasets. Our method does not rely on clients sharing the same set of classes or handling other non-local classes to cover three types of partial annotations: 1) Classes shared by all clients, 2) Classes shared by some clients, and 3) Classes unique to a client.

\subsubsection{Problem Setup} \label{sec:problem_setup} 

Let us consider a FL setup comprising of $K \in \mathbb{N}$ clients (indexed by $k$) training for $T \in \mathbb{N}$ epochs with communication after every $E \in \mathbb{N}$ epochs, resulting in $\lfloor T/E \rfloor$ communication rounds. Let $C_k$ be the set of local classes for client $k$. We define $C$ as the complete set of classes distributed across the network, such that $C = \bigcup_{k=1}^{K} C_k$.

\begin{figure}[!t]
    \centering
    \includegraphics[width = \linewidth]{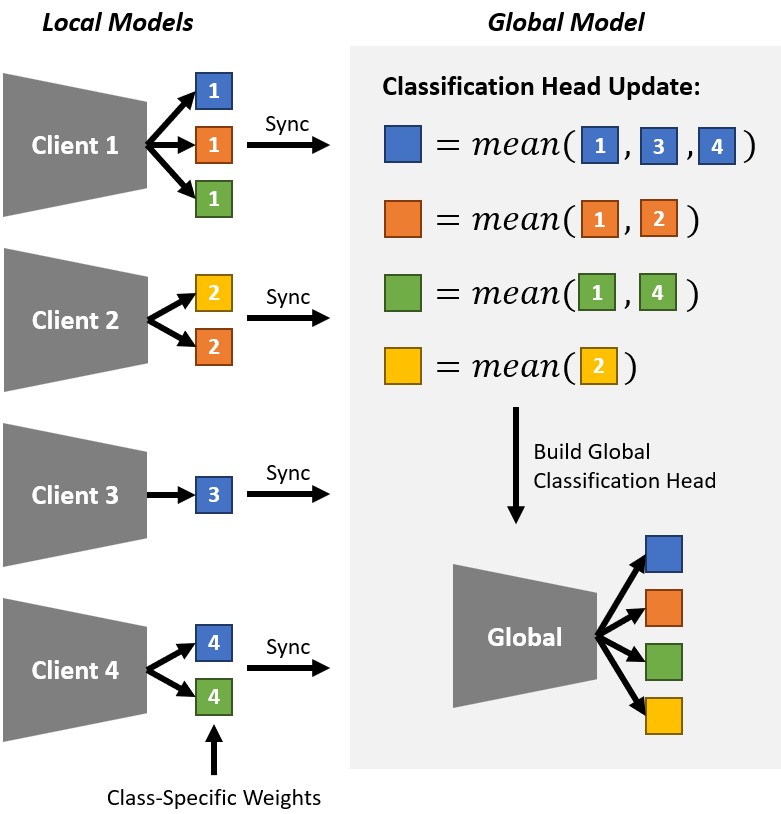}
    \caption{Illustration of surgical aggregation dynamically building a global classification head using class-specific weights.}
    \label{fig:task_build}
\end{figure}

We define a multi-label classifier $(G \circ F)$ as comprising of: 1) feature extractor $F$, and 2) classification head $G$. Here, $F$ can be any model backbone (e.g., DenseNet) and $G$ is a fully-connected layer. 

Suppose each client $k$ has local model $(G \circ F)_k$, where $F_k$ is the local feature extractor and $G_k$ is the local classification head. At an epoch $t$, we denote the weights of the local feature extractor as $\mathbf{w}_{t,k}^{(F)}$ and global feature extractor as $\mathbf{w}_{t}^{(F)}$. 

Similarly, we denote the weights of the local classification head as $\mathbf{w}_{t,k}^{(G)} \in \mathbb{R}^{N \times M_k}$ and global classification head as $\mathbf{w}_{t}^{(G)} \in \mathbb{R}^{N \times M}$. Here, $N$ is the number of input neurons to the fully-connected layer, $M$ and $M_k$ are the cardinality of $C$ and $C_k$ respectively (such that $M \geq M_k\ \forall k$). We define the class-specific weights of class $c \in C_k$ for client $k$ as column vector of $c$ in $\mathbf{w}_{t,k}^{(G)}$ and is denoted $(\mathbf{w}_{t,k}^{(G)})_c \in \mathbb{R}^{N}$. If $c \not\in C_k$, then $(\mathbf{w}_{t,k}^{(G)})_c = \mathbf{0}$.

\subsubsection{Overview of Surgical Aggregation}

At the start of each communication round, clients train locally using gradient descent for $E$ epochs on local classes before communicating the updated weights to the central server. To update the global model, we use a modified server update strategy comprising of feature extractor update and classification head update steps.

\textbf{Feature Extractor Update:} We aggregate the weights of local feature extractors $\mathbf{w}_{t,k}^{(F)}$ using model aggregation strategy $f$. Here, $f$ can be any model aggregation strategy (e.g., FedAvg). For example, if $f$ is FedAvg, then the weights of the global feature extractor are defined as \citep{mcmahan2017communication}:
\begin{equation}
\mathbf{w}_t^{(F)} = \frac{1}{K} \sum_{k=1}^{K} \mathbf{w}_{t,k}^{(l)}
\end{equation}

\begin{figure}[!t]
    \centering
    \includegraphics[width = \linewidth]{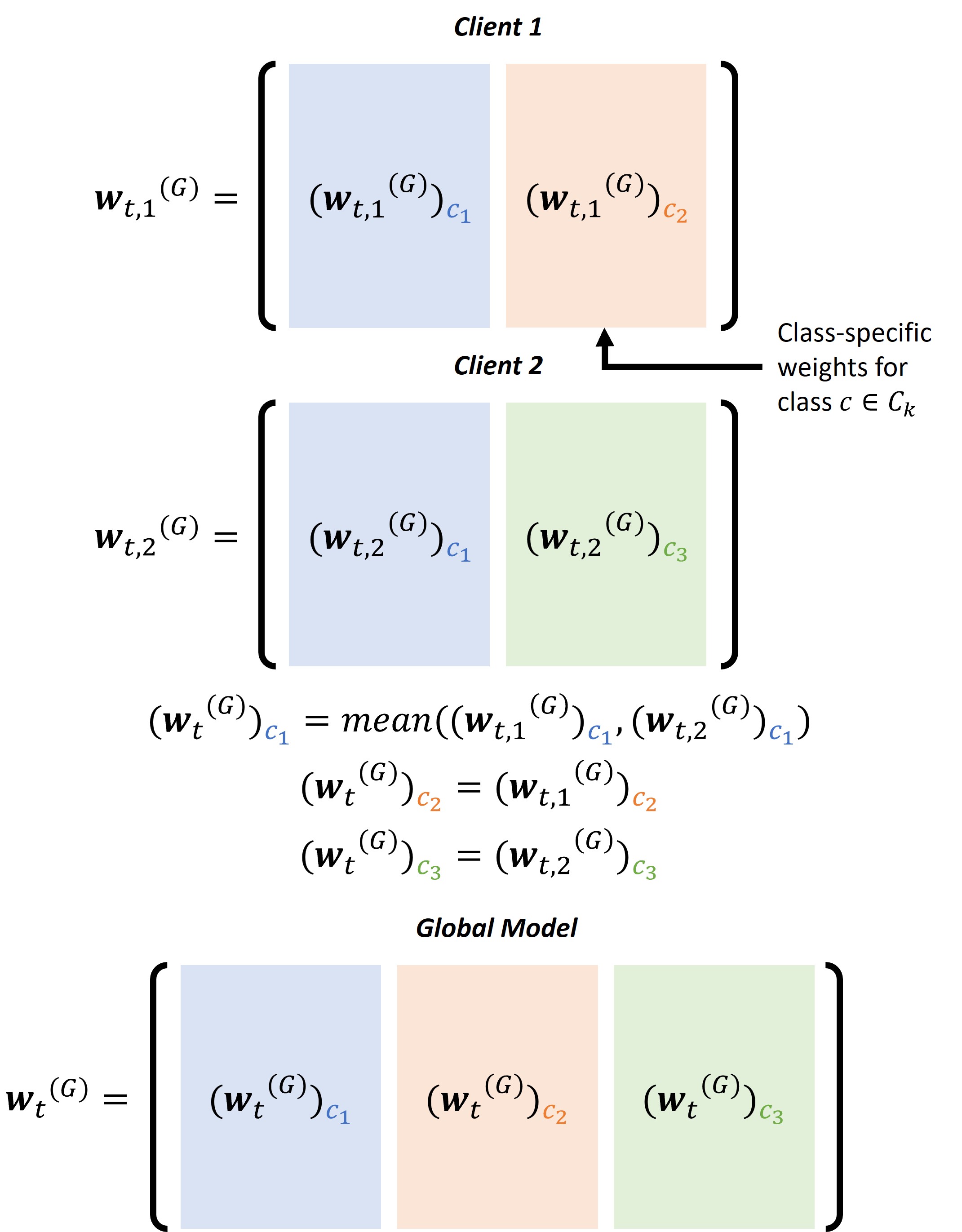}
    \caption{Detailed illustration of surgical aggregation building the weights of the global classification head.}
    \label{fig:task_build_detail}
\end{figure}

\textbf{Classification Head Update:} We selectively aggregate the class-specific weights of $c \in C$ from local classification heads $G_k$ to dynamically build the global classification head $G$, such that the global classifiers $(G \circ F)$ predicts all $C$ classes (\figureref{fig:task_build}). 

For class $c \in C$ present in $K_c \in \mathbb{N}$ clients (such that $K_c \leq K$), the global class-specific weights are computed by the mean of all local class-specific weights:
\begin{equation} \label{eq:1}
(\mathbf{w}_t^{(G)})_c = \frac{1}{K_c} \sum_{k=1}^{K} (\mathbf{w}_{t,k}^{(G)})_c
\end{equation}
If the class $c$ is unique to a client (i.e. $K_c = 1$), Equation \ref{eq:1} says that the global class-specific weights are equal to the local class-specific weights:
\begin{equation}
(\mathbf{w}_t^{(G)})_c = (\mathbf{w}_{t,k}^{(G)})_c
\end{equation}

Once the global class-specific weights are computed for all $c \in C$, the weights of the global classification head $\mathbf{w}_t$ are constructed by placing each column vector $(\mathbf{w}_t)_c$ at the index of $c$ (\figureref{fig:task_build_detail}). This enables our method to equivalent to FedAvg in the absence of class-heterogeneity. However, if class-heterogeneity is present in the network, our method employs selective aggregation to handle the weights of mismatched classification heads from clients to build a single, global classification head.

At the end of the communication round, the aggregated $\mathbf{w}_{t}^{(F)}$ is communicated back to the clients. However, our method does not directly communicate the aggregated $\mathbf{w}_{t}^{(G)}$ back to the clients because it contains class-specific weights for all classes, including non-local classes $c \not\in C_k$ for a client $k$. Instead, our method reconstructs and communicates $\mathbf{w}_{t,k}^{(G)}$ to each clients $k$, using the updated $(\mathbf{w}_{t}^{(G)})_c$ for all $c \in C_k$.

\begin{algorithm2e}[!t]
\caption{Surgical Aggregation}
\label{alg:surgagg}
\KwIn{$K$ clients indexed by $k$ with local classes $C_k$, total epochs $T$, epochs before communication $E$, initialized model parameters $\mathbf{w}_{0,k}$, model aggregation strategy $f$}
\For{$\text{each epoch}\ t \in \{1,2,...,T\}$} {
  \tcp{client update}
  \For{$\text{each client}\ k$} {
    $\mathbf{w}_{t,k} \gets SGD(\mathbf{w}_{t-1,k})$
  }
  \tcp{server update}
  \If{$\text{mod}(t, E) = 0$} {
    \tcp{feature extractor update}
    $\mathbf{w}_{t}^{(F)} \gets f(\{\mathbf{w}_{t,k}^{(F)}\ \text{: } \forall k \})$ \\

    \tcp{classification head update} 
    \For{$\text{each class}\ c \in C$} {
        $(\mathbf{w}_{t}^{(G)})_c \gets mean(\{(\mathbf{w}_{t, k}^{(G)})_c\ \text{if } c \in C_k\ \text{: } \forall k\})$ 
    }

    \tcp{communicate to clients}
    \For{$\text{each client}\ k$} {
        $\mathbf{w}_{t,k}^{(F)} \gets\mathbf{w}_{t}^{(F)}$ \\
        \For{$\text{each class}\ c \in C_k$} {
            $(\mathbf{w}_{t,k}^{(G)})_c \gets (\mathbf{w}_{t}^{(G)})_c$ 
        }
    }
  }
}
\end{algorithm2e} 

The algorithm for surgical aggregation is detailed in \algorithmref{alg:surgagg}. Convergence analysis concludes that convergence is guaranteed regardless of class-heterogeneity in the network (\appendixref{sec:convergence_analysis}). 

\subsection{Datasets}

\begin{figure}[!t]
    \centering
    \includegraphics[width = \linewidth]{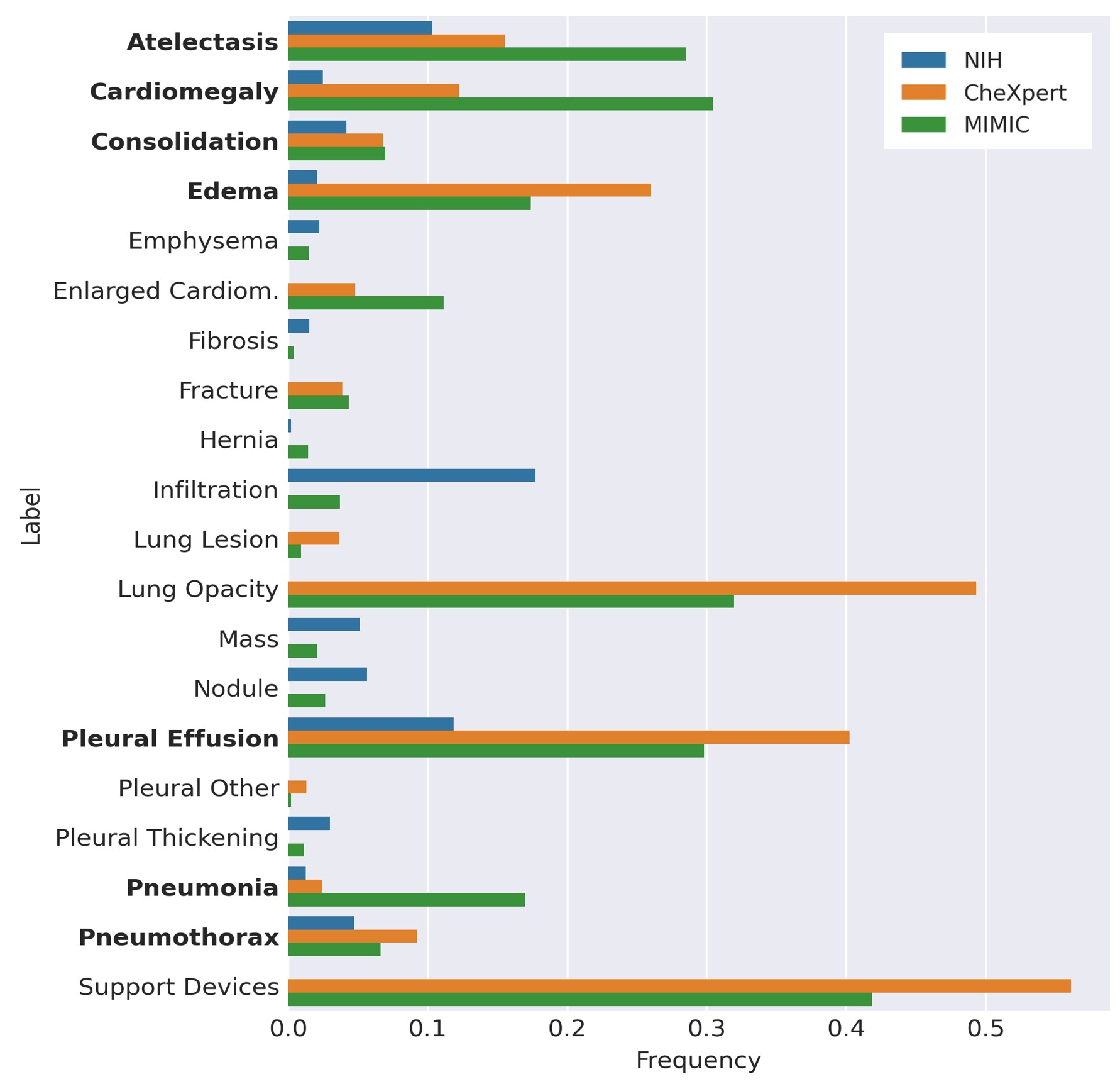}
    \caption{Freqeuence of disease labels between NIH, CheXpert, and MIMIC datasets. The seven shared classes between the NIH and CheXpert datasets are bolded.}
    \label{fig:class_heterogeneity}
\end{figure}

\begin{figure}[!t]
    \centering
    \includegraphics[width = \linewidth]{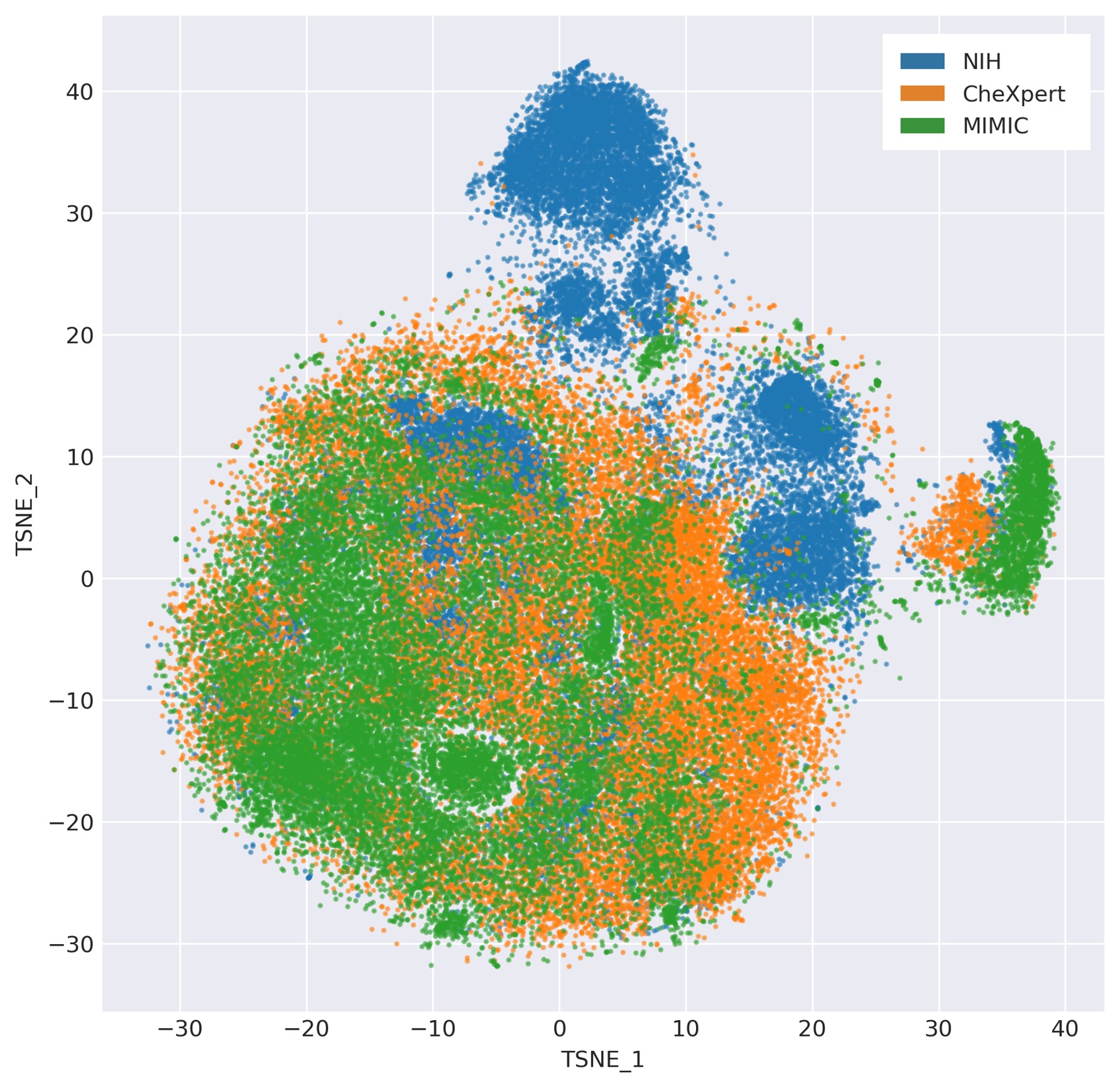}
    \caption{Latent space representation of NIH, CheXpert, and MIMIC datasets using t-SNE \citep{van2008visualizing}.}
    \label{fig:data_heterogeneity}
\end{figure}

We use three large-scale CXR datasets that are class-heterogeneous (\figureref{fig:class_heterogeneity}) and non-IID (\figureref{fig:data_heterogeneity}).

\textbf{NIH Chest X-Ray 14:} It consists of 14 disease labels with $n=112,120$ frontal CXRs from 30,805 patients \citep{wang2017chestx}. We randomly divide the dataset into training (70\%, $n=78,075$), validation (10\%, $n=11,079$), and testing (20\%, $n=22,966$) splits with no patient leakage.

\textbf{CheXpert:} It consists of 13 disease labels (seven shared with NIH) with $n=224,316$ CXRs from 65,240 patients \citep{irvin2019chexpert}. All lateral images are discarded to yield $n=191,027$ frontal CXRs. Uncertain labels are treated as negatives. We randomly divide the dataset into training (70\%, $n=133,638$), validation (10\%, $n=18,855$), and testing (20\%, $n=38,534$) splits with no patient leakage. 

\textbf{MIMIC-CXR-JPG:} It consists of $n=377,110$ CXRs from 65,379 patients and is used as our external, fully annotated test set to evaluate generalizability \citep{goldberger2000physiobank,johnson2019mimic}. All lateral images are discarded to yield $n=243,324$ frontal CXRs. The CXR-LT expanded set of 26 labels provides ground-truth for all 20 classes present in NIH and CheXpert datasets \citep{holste2023does}.

\subsection{Baselines}

We compare our method to two baselines that represent an upper bound of expected model performance, but do not yield a singular global model.

\textbf{Individually Trained:} Models trained individually on each dataset in a centralized setup using standard DL.
    
\textbf{Personalized FL:} PFL is a popular approach to FL, where each client learns a personalized local model in favor of generalized global models \citep{liang2020think,tan2022towards,li2021fedbn}. While clients share a feature extractor, they learn a personalized classification head for local classes.

We also compare our method to three existing methods for training models with class-heterogeneous datasets that result in a single global model.

\textbf{Centralized:} Multiple datasets are concatenated locally to train a model in a centralized setup using standard DL to achieve better generalizability \citep{cohen2020limits,seyyed2021underdiagnosis}. All missing classes in the combined dataset are treated as negatives. For a client $k$ with local classes $C_k$, and data $\{\mathbf{x}_k, y_k\}$ containing $n_k$ samples (indexed by $i$), $(y_{k,i})_c = 0$ for all $c \not\in C_k$.

\textbf{Vanilla FL:} Conventional approach for FL where datasets are distributed across clients in the network \citep{chowdhury2022review,sheller2020federated}. Missing classes in each dataset are treated as negatives.

\textbf{FL with Partial Loss:} A modification of the vanilla FL setup using a partial loss function that drops non-local classes when computing local losses \citep{jin2002learning}. For client $k$ with model parameters $\mathbf{w}_{t,k}$, local classes $C_k$, and private data $\{\mathbf{x}_k, y_k\}$ containing $n_k$ samples (indexed by $i$), it is defined as:
\begin{equation}
    L(\mathbf{w}_{t,k}, \mathbf{x}_{k,i}, y_{k,i}) = \frac{1}{|C_k|} \sum_{c \in C_k} \ell(\mathbf{w}_{t,k}, \mathbf{x}_{k,i}, (y_{k,i})_c)
\end{equation}

\subsection{Experiments} \label{sec:experiments}

\subsubsection{Ablation Study}

In this experiment, we evaluated the effect of 1) number of clients and 2) number of shared classes using simulated datasets. We sampled IID subsets of the NIH dataset to create simulated datasets for each client, such that there is no patient leakage and all 14 classes from the NIH dataset appear at least once in the network. Since each simulated dataset is an "incomplete" snapshot of the NIH dataset, we evaluated the performance of surgical aggregation in relation to the NIH baseline model.

\textbf{Effect of number of clients:} We evaluated the effect of number of clients across seven iterations, each with an increasing number of clients $K \in \{2,3,4,5,6,8,10\}$. We randomly selected values of $K$, such that as $K$ increases, the sample size of the simulated dataset at each client $k$ decreases by factor of $1/K$.

\textbf{Effect of number of shared classes:} We evaluated the effect of number of shared classes as a proxy for class-heterogeneity in the network. Since the NIH dataset contains 14 classes, we randomly sampled seven possible iterations out of 14, each with an increasing number of shared classes $\{0,1,2,4,8,12,14\}$. The number of clients $K=4$ and underlying data at each client was constant.

All models were evaluated on the held-out NIH test set and the external MIMIC dataset. We used FedBN+ for all FL approaches \citep{kulkarni2023fedbn}. The PFL baseline was not included because it does not train a global model.

\subsubsection{Non-IID Experiments}

We evaluated surgical aggregation using the NIH and CheXpert datasets across two sub-experiments:

\textbf{Non-Class-Heterogeneous:} We considered a purely non-IID setting with no class-heterogeneity, where only the seven shared classes between both datasets (see \figureref{fig:class_heterogeneity}) were included as ground-truth annotations. All models were evaluated on the NIH and CheXpert test sets and the external MIMIC dataset using the seven shared classes. 

\textbf{Class-Heterogeneous:} We considered a non-IID and class-heterogeneous setting, where all 20 classes from both datasets were included as ground-truth annotations. All models were evaluated on the NIH and CheXpert test sets using local classes (14 and 13 respectively) and the external MIMIC dataset using all 20 classes. We also measured the disparity in performance between the shared and unique classes for the NIH and CheXpert datasets using their respective test set. Furthermore, we measured the performance on the NIH and CheXpert classes when evaluating with the external MIMIC test set. This enabled us to compare our method with the individually trained and PFL baselines.

\subsection{Implementation}

\textbf{Centralized Setup:} We trained ImageNet-pretrained DenseNet121 models using binary cross-entropy (BCE) loss and a batch size of 64. Prior to training, the classification heads were "warmed up" using transfer learning for 15 epochs with a learning rate of 5e-3, while keeping feature extractor weights frozen. Then, the model was trained with all weights unfrozen for 150 epochs with a learning rate of 5e-5.

\textbf{Federated Setup:} We simulated a FL setup with $T=150$ epochs and $E=1$ epochs before communication. Each client trained an ImageNet-pretrained DenseNet121 model using the methods described above. The classification heads of local models were similarly "warmed up" prior to training. Global convergence was determined by the mean of local validation BCE loss for each client. We considered the three aggregation strategies: 1) FedAvg, the de-facto FL strategy \citep{mcmahan2017communication}. 2) FedBN, a variation of FedAvg with local batch normalization \citep{li2021fedbn}. As a PFL strategy, FedBN is only included for the PFL baseline model. 3) FedBN+, a modification of FedBN with pre-trained batch normalization statistics for global models \citep{kulkarni2023fedbn}.

In order to primarily focus on class-heterogeneity, we used the same hyperparameters for all models trained. All CXRs were downsampled to $224 \times 224$, normalized between 0 and 1, and scaled to ImageNet statistics. Random augmentations (rotation, flip, zoom, and contrast) were applied to input images during training. All models were trained and evaluated using TensorFlow (version 2.8.1) and CUDA (version 12.0) on four NVIDIA RTX A6000 GPUs.

\begin{figure*}[!ht]
    \centering
    \includegraphics[width = \linewidth]{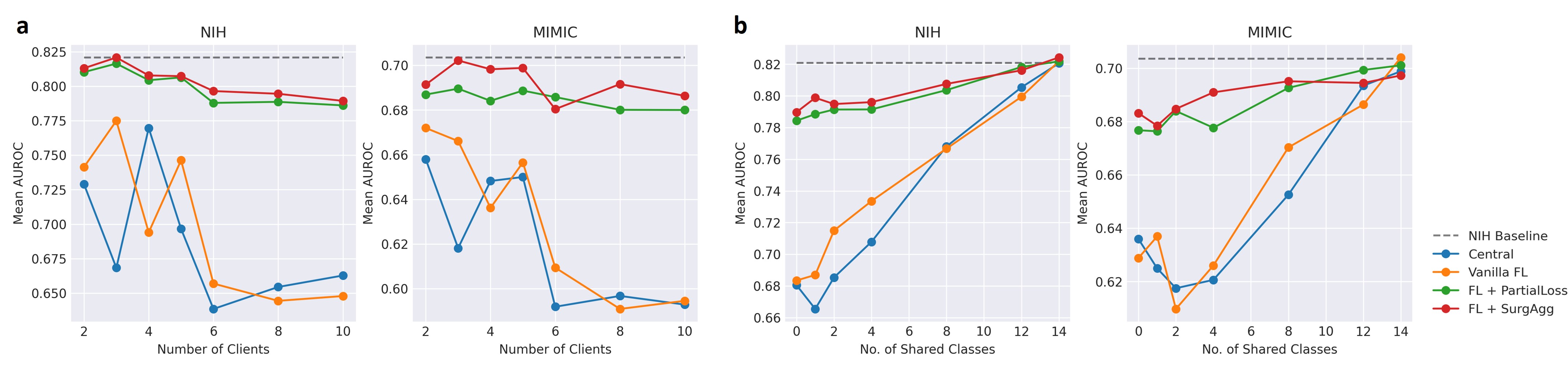}
    \caption{Mean AUROC scores for ablation study evaluating \textbf{(a)} effect of number of clients, \textbf{(b)} effect of number of shared classes on the NIH (left) and MIMIC (right) test sets.}
    \label{fig:exp1_results}
\end{figure*}

\begin{figure}[!t]
    \centering
    \includegraphics[width = 0.8\linewidth]{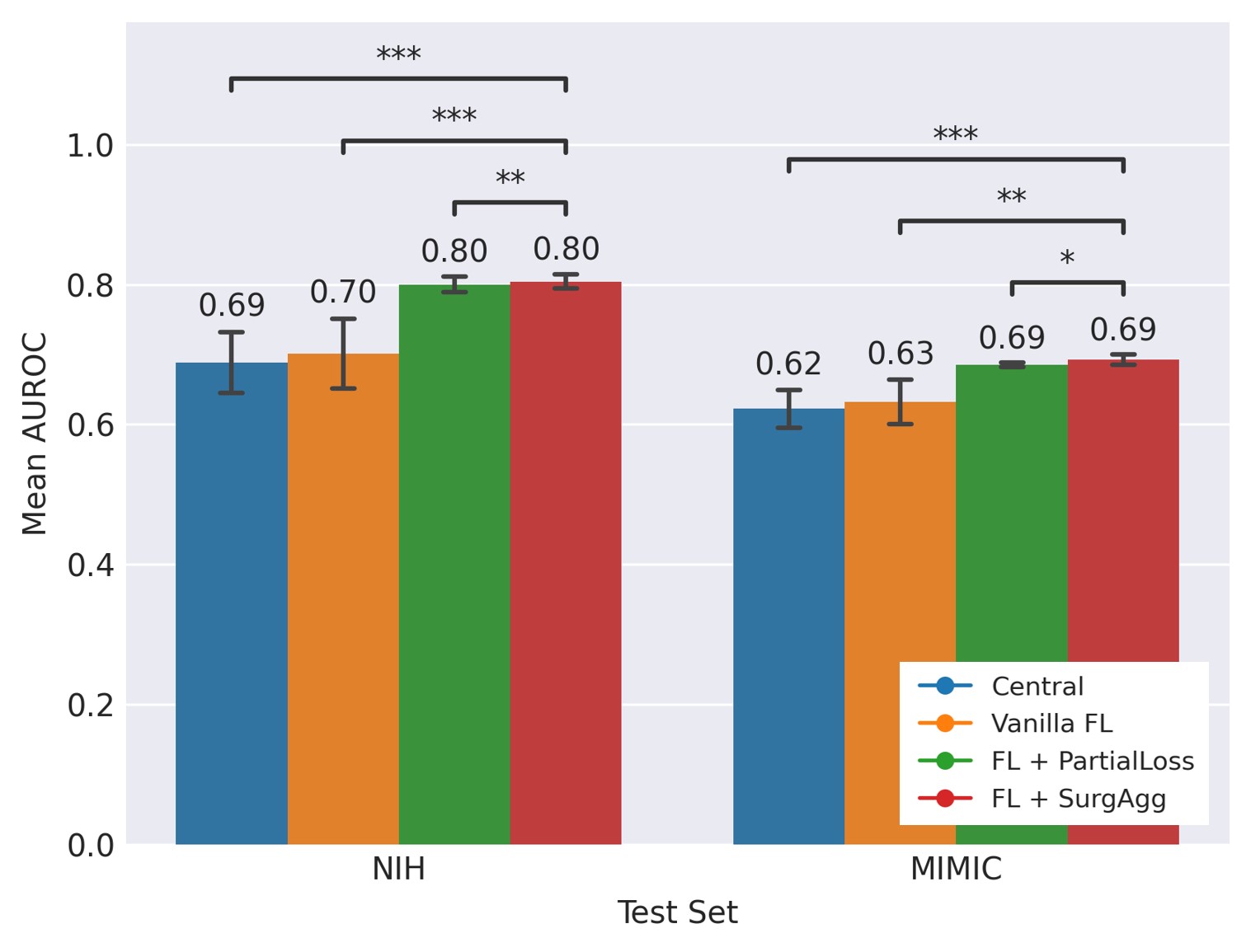}
    \caption{Average mean AUROC scores for the effect of number of clients on NIH and MIMIC test sets. (ns: $p>0.05$, *: $p\leq0.05$, **: $p\leq0.01$, ***:  $p\leq0.001$)}
    \label{fig:exp1.1_mean_results}
\end{figure}

\begin{figure}[!t]
    \centering
    \includegraphics[width = 0.8\linewidth]{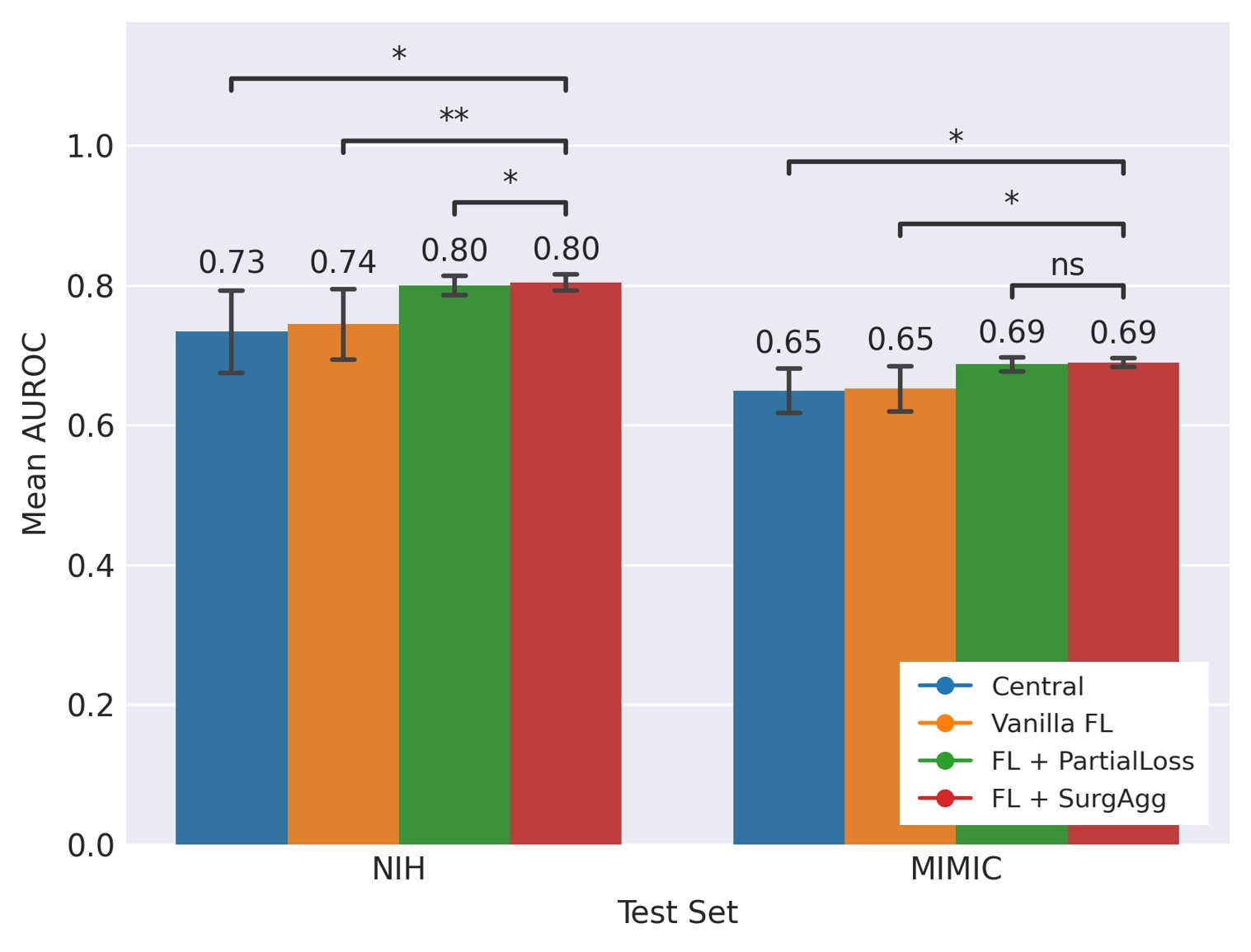}
    \caption{Average mean AUROC scores for the effect of number of shared classes on NIH and MIMIC test sets. (ns: $p>0.05$, *: $p\leq0.05$, **: $p\leq0.01$, ***:  $p\leq0.001$)}
    \label{fig:exp1.2_mean_results}
\end{figure}

\subsection{Metrics and Statistical Analyses} \label{sec:metrics}

Model performance in a multi-label scenario is quantified by the mean of area under the receiver operating characteristic curve (AUROC) scores for each class in the test set. If a class in the test set is absent from the set of classes learned by a model, the mean AUROC score is not defined. All values are reported as Mean $\pm$ SD.

For the ablation study, performance between models is compared using paired t-tests between the mean AUROC scores to test for significance. For real-world experiments, performance between models is compared using paired t-tests across the granular class-level AUROC scores to test for significance. In all cases, statistical significance was defined as $p < 0.05$. 

\section{Results} \label{sec:results}

\subsection{Ablation Study}

\textbf{Effect of number of clients:} We observe that surgical aggregation outperforms all existing methods with average mean AUROC $0.80 \pm 0.01$ on the held-out NIH test set (all $p\leq0.005$) (Figures \ref{fig:exp1_results}(a) and \ref{fig:exp1.1_mean_results}). While FL with partial loss performs comparably to our method, the difference in average mean AUROCs ($\Delta=0.004$) is statistically significant ($p=0.004$). On the external MIMIC dataset, our method generalizes better with average mean AUROC $0.69 \pm 0.01$, compared to all existing methods (all $p\leq0.03$).

\textbf{Effect of number of shared clients:} We observe that surgical aggregation outperforms all existing methods with average mean AUROC $0.80 \pm 0.01$ on the held-out NIH test set (all $p\leq0.03$) (Figures \ref{fig:exp1_results}(b) and \ref{fig:exp1.2_mean_results}). Similarly, the minor differences in average mean AUROCs between FL with partial loss and surgical aggregation are statistically significant ($\Delta=0.004$, $p=0.03$). On the external MIMIC dataset, our method generalizes better than all existing methods with average mean AUROC $0.69 \pm 0.01$.

Finally, our results indicate that even with varying levels of class- and data-heterogeneity in the network, surgical aggregation provides the closest approximation to the performance of the NIH baseline model (mean AUROC $0.82$, mean $\Delta=0.017$), followed by FL with partial loss (mean $\Delta=0.021$).

\begin{table}[!t]
    \centering
    \caption{Mean AUROC scores on NIH, CheXpert, and MIMIC dataset for non-IID, non-class-heterogeneous experiment. Our method is compared with others using paired t-tests. Best performing method is bolded.}
    \label{tab:exp2a_results}
    \footnotesize
    \begin{tabular*}{\linewidth}{@{\extracolsep{\fill}} lccc}
          & \multicolumn{1}{c}{\textbf{NIH}} & \multicolumn{1}{c}{\textbf{CheXpert}} & \multicolumn{1}{c}{\textbf{MIMIC}} \\ \toprule
        NIH & $0.84 \pm 0.05^{\text{ns}}$ & $0.70 \pm 0.07^{**}$ & $0.73 \pm 0.10^{**}$ \\
        CheXpert & $0.81 \pm 0.06^{*}$ & $0.76 \pm 0.07^{\text{ns}}$ & $0.75 \pm 0.09^{\text{ns}}$ \\
        PFL \\
        \quad FedAvg & $0.83 \pm 0.06^{*}$ & $0.76 \pm 0.07^{\text{ns}}$ & $0.76 \pm 0.09^{\text{ns}}$ \\
        \quad FedBN & $0.83 \pm 0.05^{*}$ & $0.77 \pm 0.07^{\text{ns}}$ & $0.75 \pm 0.08^{\text{ns}}$ \\
        \quad FedBN+ & $0.83 \pm 0.06^{**}$ & $0.76 \pm 0.07^{\text{ns}}$ & $0.75 \pm 0.08^{\text{ns}}$ \\
        \midrule
        Central & $0.82 \pm 0.06^{**}$ & $0.77 \pm 0.07^{\text{ns}}$ & $0.76 \pm 0.10^{\text{ns}}$ \\
        FL \\
        \quad FedAvg & $0.81 \pm 0.05^{***}$ & $0.74 \pm 0.08^{*}$ & $0.74 \pm 0.10^{*}$ \\
        \quad FedBN+ & $0.84 \pm 0.05^{*}$ & $0.75 \pm 0.08^{*}$ & $0.75 \pm 0.09^{\text{ns}}$ \\
        FL + PL \\
        \quad FedAvg & $0.82 \pm 0.05^{***}$ & $0.75 \pm 0.08^{*}$ & $0.74 \pm 0.10^{*}$ \\
        \quad FedBN+ & $0.83 \pm 0.06^{*}$ & $0.75 \pm 0.07^{\text{ns}}$ & $0.76 \pm 0.09^{\text{ns}}$ \\
        \textbf{FL + SA} \\
        \quad FedAvg & $0.84 \pm 0.05^{*}$ & $0.75 \pm 0.07^{\text{ns}}$ & $0.75 \pm 0.09^{\text{ns}}$ \\
        \quad FedBN+ & $\mathbf{0.84} \pm 0.06$ & $\mathbf{0.77} \pm 0.08$ & $\mathbf{0.76} \pm 0.10$ \\ \bottomrule
        \multicolumn{4}{l}{PL: Partial Loss, SA: Surgical Aggregation} \\
        \multicolumn{4}{l}{(ns: $p>0.05$, *: $p\leq0.05$, **: $p\leq0.01$, ***: $p\leq0.001$)}
    \end{tabular*}
\end{table}

\subsection{Non-IID Experiments}

\textbf{Non-Class-Heterogeneous:} We observe that the non-class-heterogeneous experiment results in similar mean AUROCs across the held-out NIH and CheXpert test sets and the external MIMIC dataset. This suggests that surgical aggregation is comparable to current FL approaches in the absence of class-heterogeneity (\tableref{tab:exp2a_results}). 

On the NIH test set, our method performs comparably to the NIH baseline with mean AUROC $0.84 \pm 0.06$ ($p=0.22$), while outperforming other baselines (all $p\leq0.03$) and existing methods (all $p\leq0.02$). On the CheXpert test set, our method performs comparably to the CheXpert baseline model with mean AUROC $0.77 \pm 0.08$ ($p=0.51$). Our method performs on par with the PFL baselines (all $p>0.05$) and existing methods, while outperforming the NIH baseline ($p=0.002$)

Finally, on the external MIMIC dataset, we observe that surgical aggregation's generalizability to unseen data is on par with baselines with mean AUROC $0.76 \pm 0.10$ (all $p>0.05$), while outperforming the NIH baseline ($p=0.009$). Similarly, our method performs comparably to the existing methods.

\begin{table}[!t]
    \centering
    \caption{Mean AUROC scores on NIH, CheXpert and MIMIC test sets for non-IID, class-heterogeneous experiment. Our method is compared with others using paired t-tests. Best performing method is bolded.}
    \label{tab:exp2b_results}
    \footnotesize
    \begin{tabular*}{\linewidth}{@{\extracolsep{\fill}} lccc}
          & \multicolumn{1}{c}{\textbf{NIH}} & \multicolumn{1}{c}{\textbf{CheXpert}} & \multicolumn{1}{c}{\textbf{MIMIC}} \\ \toprule
        NIH & $0.82 \pm 0.06^{\text{ns}}$ & - & - \\
        CheXpert & - & $0.74 \pm 0.07^{*}$ & - \\
        PFL \\
        \quad FedAvg & $0.79 \pm 0.07^{***}$ & $0.73 \pm 0.07^{***}$ & - \\
        \quad FedBN & $0.81 \pm 0.06^{\text{ns}}$ & $0.74 \pm 0.06^{*}$ & - \\
        \quad FedBN+ & $0.81 \pm 0.06^{\text{ns}}$ & $0.73 \pm 0.07^{*}$ & - \\
        \midrule
        Central & $0.71 \pm 0.13^{**}$ & $0.72 \pm 0.07^{***}$ &
        $0.65 \pm 0.10^{***}$ \\
        FL \\
        \quad FedAvg & $0.70 \pm 0.10^{***}$ & $0.68 \pm 0.08^{***}$ & $0.64 \pm 0.09^{***}$ \\
        \quad FedBN+ & $0.71 \pm 0.12^{***}$ & $0.68 \pm 0.09^{***}$ & $0.66 \pm 0.10^{***}$ \\
        FL + PL \\
        \quad FedAvg & $0.77 \pm 0.09^{***}$ & $0.68 \pm 0.07^{***}$ & $0.68 \pm 0.09^{***}$ \\
        \quad FedBN+ & $0.78 \pm 0.07^{***}$ & $0.71 \pm 0.07^{***}$ & $0.68 \pm 0.09^{***}$ \\
        \textbf{FL + SA} \\
        \quad FedAvg & $0.80 \pm 0.06^{*}$ & $0.73 \pm 0.07^{**}$ & $0.70 \pm 0.09^{\text{ns}}$ \\
        \quad FedBN+ & $\mathbf{0.81} \pm 0.07$ & $\mathbf{0.76} \pm 0.08$ & $\mathbf{0.71} \pm 0.10$ \\ \bottomrule
        \multicolumn{4}{l}{PL: Partial Loss, SA: Surgical Aggregation} \\
        \multicolumn{4}{l}{(ns: $p>0.05$, *: $p\leq0.05$, **: $p\leq0.01$, ***: $p\leq0.001$)}
    \end{tabular*}
\end{table}

\textbf{Class-Heterogeneous:} We observe that class-heterogeneity results in a severe degradation of model performance for the existing methods when compared to the non-class-heterogeneous experiment. However, the baseline methods and our method continue to demonstrate high-performance (\tableref{tab:exp2b_results}). 

\begin{table*}[!t]
    \centering
    \caption{Sub-analysis of mean AUROC scores for shared and unique classes of the NIH and CheXpert datasets for non-IID, class-heterogeneous experiment. Our method is compared with others using paired t-tests. Best performing method is bolded.}
    \label{tab:exp2b_results_subanalysis}
    \footnotesize
    \begin{tabular*}{0.8\linewidth}{@{\extracolsep{\fill}} lcccc}
          & \multicolumn{2}{c}{\textbf{NIH}} & \multicolumn{2}{c}{\textbf{CheXpert}} \\ 
        \cmidrule(lr){2-3} \cmidrule(lr){4-5}
          & \multicolumn{1}{c}{\textbf{Shared Classes}} & \multicolumn{1}{c}{\textbf{Unique Classes}} & \multicolumn{1}{c}{\textbf{Shared Classes}} & \multicolumn{1}{c}{\textbf{Unique Classes}} \\
        \toprule
        NIH & $0.84 \pm 0.06^{\text{ns}}$ & $0.80 \pm 0.07^{\text{ns}}$ & $0.70 \pm 0.08^{**}$ & - \\
        CheXpert & $0.81 \pm 0.07^{*}$ & - & $0.76 \pm 0.07^{\text{ns}}$ & $0.71 \pm 0.05^{**}$ \\
        PFL & $0.83 \pm 0.06^{**}$ & $0.80 \pm 0.07^{\text{ns}}$ & $0.76 \pm 0.07^{\text{ns}}$ & $0.70 \pm 0.05^{**}$ \\
        \midrule
        Central & $0.82 \pm 0.05^{*}$ & $0.60 \pm 0.07^{***}$ & $0.74 \pm 0.07^{\text{ns}}$ & $0.69 \pm 0.06^{**}$ \\
        FL & $0.82 \pm 0.06^{**}$ & $0.60 \pm 0.06^{***}$ & $0.73 \pm 0.09^{**}$ & $0.62 \pm 0.05^{**}$ \\
        FL + PL & $0.82 \pm 0.05^{*}$ & $0.73 \pm 0.06^{***}$ & $0.74 \pm 0.07^{*}$ & $0.67 \pm 0.04^{**}$ \\
        \textbf{FL + SA} & $\mathbf{0.84} \pm 0.06$ & $\mathbf{0.78} \pm 0.07$ & $\mathbf{0.77} \pm 0.09$ & $\mathbf{0.75} \pm 0.07$ \\ \bottomrule
        \multicolumn{5}{l}{PL: Partial Loss, SA: Surgical Aggregation} \\
        \multicolumn{5}{l}{(ns: $p>0.05$, *: $p\leq0.05$, **: $p\leq0.01$, ***: $p\leq0.001$)}
    \end{tabular*}
\end{table*}

\begin{table}[!t]
    \centering
    \caption{Sub-analysis of mean AUROC scores for NIH and CheXpert classes on the MIMIC test for non-IID, class-heterogeneous experiment. Our method is compared with others using paired t-tests. Best performing method is bolded.}
    \label{tab:exp2b_mimic_results}
    \footnotesize
    \begin{tabular*}{\linewidth}{@{\extracolsep{\fill}} lcc} 
          & \multicolumn{2}{c}{\textbf{MIMIC}} \\ 
          \cmidrule(lr){2-3}
          & \multicolumn{1}{c}{\textbf{NIH Classes}} & \multicolumn{1}{c}{\textbf{CheXpert Classes}} \\ \toprule
        NIH & $0.73 \pm 0.10^{\text{**}}$ & -  \\
        CheXpert & - & $0.67 \pm 0.11^{*}$ \\
        PFL \\
        \quad FedAvg & $0.74 \pm 0.11^{***}$ & $0.69 \pm 0.10^{\text{ns}}$ \\
        \quad FedBN & $0.70 \pm 0.09^{***}$ & $0.67 \pm 0.10^{\text{ns}}$ \\
        \quad FedBN+ & $0.74 \pm 0.10^{**}$ & $0.68 \pm 0.11^{\text{ns}}$ \\
        \midrule
        Central & $0.74 \pm 0.09^{**}$ & $0.66 \pm 0.11^{**}$ \\
        FL \\
        \quad FedAvg & $0.72 \pm 0.10^{***}$ & $0.62 \pm 0.06^{**}$ \\
        \quad FedBN+ & $0.75 \pm 0.10^{**}$ & $0.63 \pm 0.08^{*}$ \\
        FL + PL \\
        \quad FedAvg & $0.74 \pm 0.09^{**}$ & $0.66 \pm 0.09^{*}$ \\
        \quad FedBN+ & $0.74 \pm 0.10^{**}$ & $0.64 \pm 0.10^{*}$ \\
        \textbf{FL + SA} \\
        \quad FedAvg & $0.75 \pm 0.09^{\text{ns}}$ & $0.68 \pm 0.10^{\text{ns}}$ \\
        \quad FedBN+ & $\mathbf{0.76} \pm 0.10$ & $\mathbf{0.70} \pm 0.12$ \\ \bottomrule
        \multicolumn{3}{l}{PL: Partial Loss, SA: Surgical Aggregation} \\
        \multicolumn{3}{l}{(ns: $p>0.05$, *: $p\leq0.05$, **: $p\leq0.01$, ***: $p\leq0.001$)}
    \end{tabular*}
\end{table}

On the NIH test set, surgical aggregation performs comparably to the baseline models with mean AUROC $0.81 \pm 0.07$ (all $p>0.05$), apart from the FedAvg PFL baseline ($p\leq0.001$). Furthermore, our method outperforms the existing methods (all $p\leq0.002$). On the CheXpert test set, surgical aggregation outperforms all baselines (all $p\leq0.03$) and existing methods (all $p\leq0.004$) with mean AUROC $0.76 \pm 0.08$. Finally, on the external MIMIC dataset, surgical aggregation outperforms all existing methods with mean AUROC $0.7 1\pm 0.10$ (all $p\leq0.001$).

When exploring the mean AUROC scores of shared and unique classes in the NIH and CheXpert datasets, we observe that the existing methods perform poorly on the unique classes when compared to the baselines while our method demonstrates high-performance on the unique classes (\tableref{tab:exp2b_results_subanalysis}). On the NIH test set, our method performs on par with the baselines ($p>0.05$), but on the CheXpert test set, surgical aggregation performs comparably to the baselines on the shared classes (both $p>0.05$), while outperforming them on the unique classes (both $p\leq0.008$).

When the mean AUROC scores of the NIH and CheXpert classes is measured separately for the MIMIC test set, we observe that surgical aggregation generalizes better than the baselines and existing methods on NIH classes with mean AUROC $0.76 \pm 0.10$ (all $p\leq0.006$) (\ref{tab:exp2b_mimic_results}). Similarly, on CheXpert classes, our method yields mean AUROC $0.70 \pm 0.12$ and outperforms the existing methods (all $p\leq0.02$), while performing comparably to the baselines (all $p>0.05$), except the CheXpert baseline ($p=0.04$).

\section{Discussion} \label{sec:discussion}

Class-heterogeneity is a significant challenge in FL, with most FL methods relying on assumptions that all clients either share the same classes, know the classes of other clients in the network, or have access to a fully annotated public dataset. However, there is a significant gap in literature for methods that address class-heterogeneity. Our results show that surgical aggregation can train a global model using distributed, class-heterogeneous datasets.

In the absence of class-heterogeneity, our method performs on par with well-established approaches in literature. However, in the presence of class-heterogeneity, our method continues to maintain high-performance and outperforms the baselines. It also scales across both IID and non-IID settings. On the other hand, existing methods fall short and scale poorly in non-IID settings. For example, while FL with partial loss is a close second in performance with IID data, its performance significantly degrades in non-IID settings. This is due to feature shift as local updates from a client for non-local classes leads to client drift and performance degradation on non-shared classes \citep{li2021fedbn,cohen2020limits}. Unlike our method, FL with partial loss requires clients to account for non-local classes and does not support heterogeneous local classification heads. Therefore, selective aggregation is critical to solve class-heterogeneity in non-IID settings.

Another advantage of our method is its ability to learn a global model that can predict all classes observed in the network. While it is popular approach in non-IID settings, PFL entails $K$ clients learning $K$ personalized models. This has interesting implications in medical imaging: 1) Each personalized model will yield different predictions for class shared between two clients, thus, raising the question of which model's prediction should be trusted. 2) PFL emphasizes local optimization over generalization, thus making personalized models susceptible to poor generalizability \citep{zhang2023grace}.

These advantages enable surgical aggregation to not just facilitate greater collaboration between institutes focusing on different tasks in the same domain, but also enable the countless "narrowly" focused datasets (e.g., those curated for competitions) to be leveraged to collectively train models with a complete set of abnormalities.

Despite our results demonstrating the advantages of surgical aggregation, our work has certain limitations: 1) There are no established benchmarks for class-heterogeneity beyond external validation. 2) Our method requires clients to share the feature extractor, an important consideration for device heterogeneity. 3) Our experiments were conducted on a simulated FL with a fixed $E=1$ epochs before communication. 4) Our experiments lacked cross-validation and the initial split of data could influence our method's performance. 5) Due to the primary focus on class-heterogeneity, our ablation study did not evaluate the effect of class-imbalance or non-IID data. For future work, we intend to explore class-heterogeneity using real-world FL setups (e.g., Flower \citep{beutel2020flower}) and further validate our method across different domains and tasks (such as 3D segmentation).

\bibliography{references}

\appendix
\counterwithin{figure}{section}
\counterwithin{equation}{section}

\newpage
\section{Convergence Analysis} \label{sec:convergence_analysis}

Consider the problem setup from \sectionref{sec:problem_setup}. Most FL approaches operate under the constraint the all clients share the same set of classes, i.e., $\forall c \in C_i$, $c \in C_j\ \forall i \neq j$. However, class-heterogeneity necessitates a broader definition:

\textbf{Definition 1.} \label{def:2} \textit{We define class-heterogeneity as the case that covers: 1) Classes shared by all clients in the network, i.e., $\exists c \in C_i$ such that $c \in C_j\ \forall i \neq j$, 2) Classes partially shared by some clients in the network, i.e., $\exists c \in C_i$ such that $c \in C_j\ \exists i \neq j$, and 3) Classes unique to a client in the network, i.e., $\exists c \in C_i$ such that $c \not\in C_j\ \forall i \neq j$.}

Let us consider that each client $k$ contains private data $\{\mathbf{x}_k, y_k\}$ containing $n_k$ samples (indexed by $i$) and is initiated with parameters $\mathbf{w}_{0,k}$ updated using gradient descent.

We primarily focus this analysis on the convergence of selective aggregation to dynamically build the global classification head. In this case, our distributed optimization problem is defined as:
\begin{equation}
    \min_{\mathbf{w}} \left\{ (G \circ F)^*(\mathbf{w}) \triangleq \frac{1}{K}\sum_{k=1}^{K} (G \circ F)_k^*(\mathbf{w}) \right\}
\end{equation}
where the local objective $(G \circ F)_k^*$ is defined by:
\begin{equation}
    (G \circ F)_k^* \triangleq \frac{1}{n_k} \sum_{i=1}^{n_k} L(\mathbf{w}, \mathbf{x}_{k,i}, y_{k,i})
\end{equation}
where $L$ is the loss function (e.g., binary cross-entropy) \citep{li2019convergence}.

\textbf{Lemma 1.} \label{lem:1} \textit{The optimization of the local objective $(G \circ F)^*_k$ for client $k$ is controlled by the optimization of $(G \circ F)^*_k$ on each individual class $c \in C_k$.}

\textbf{Proof:} For multi-label classifiers, the loss at client $k$ is computed as the mean loss for each class $c \in C_k$ using loss function $L$, such that:
\begin{equation}
    L(\mathbf{w}, \mathbf{x}_{k,i}, y_{k,i}) = \frac{1}{|C_k|} \sum_{c \in C_k} \ell(\mathbf{w}, \mathbf{x}_{k,i}, (y_{k,i})_c)
\end{equation}
where $(y_{k,i})_c$ is the ground truth for class $c$ and sample $i \in [n_k]$ at client $k$. Therefore, to minimize $L(\mathbf{w}, \mathbf{x}_{k,i}, y_{k,i})$, $\ell(\mathbf{w}, \mathbf{x}_{k,i}, (y_{k,i})_c)$ must be minimized for all $c \in C_k$. In other words, optimizing $(G \circ F)^*_k$ on each individual class $c \in C_k$ ensures optimization of the local objective $(G \circ F)^*_k$.

Furthermore, the global objective $(G \circ F)^*$ can be interpreted as the optimization of two concurrent sub-objectives:
\begin{enumerate}
    \item Optimizing the global feature extractor $F$ by locally optimizing client feature extractors $F_k$, such that:
    \begin{equation}
        \min_{\mathbf{w}} \left\{ F^*(\mathbf{w})  \triangleq \frac{1}{K}\sum_{k=1}^{K} F_k^*(\mathbf{w}) \right\}
    \end{equation}

    \item Optimizing the global classifier $G$ by locally optimizing client classifiers $G_k$, such that:
    \begin{equation}
        \min_{\mathbf{w}} \left\{ G^*(\mathbf{w})  \triangleq \frac{1}{K}\sum_{k=1}^{K} G_k^*(\mathbf{w}) \right\}
    \end{equation}
\end{enumerate}
where the local sub-objectives $F_k^* = G_k^* \triangleq (G \circ F)_k^*$.

\textbf{Lemma 2.} \label{lem:2} \textit{The optimization of the global sub-objective $F^*$ is independent of class-heterogeneity present in the network.}

\textbf{Proof:} Like conventional FL approaches, in our approach the feature extractor weights $\mathbf{w}_{t,k}^{(F)}$ are always shared with all clients at the end of each communication round, such that:
\begin{align}
    \mathbf{w}_{t}^{(F)} &= f(\{\mathbf{w}_{t,k}^{(F)}\ \text{: } \forall k \}) \\
     \mathbf{w}_{t,k}^{(F)} &=  \mathbf{w}_{t}^{(F)}\ \forall k
\end{align}
In this case, the convergence of $\mathbf{w}_{t,k}^{(F)}$ is well-established in prior literature \citep{li2019convergence,li2021fedbn}. However, it is not clear whether class-heterogeneity affects the convergence of $\mathbf{w}_{t,k}^{(F)}$. The optimization of the global sub-objective $F^*$ is controlled by the optimization of the local objective $(G \circ F)^*_k$. Therefore, the optimization of $F^*$ is ensured if $F_k^* \triangleq (G \circ F)^*_k$ is optimized. While the optimization of $(G \circ F)^*_k$ is dependent on classes $C_k$, it is independent of the optimization of classes $c \in C \backslash C_k$ (Lemma \hyperref[lem:1]{1}). In other words, the optimization of $F^*$ is driven by the optimization of $F_k^*$, regardless of any class-heterogeneity.

Since the classifier weights are selectively aggregated, we analyze the convergence of the global classifier (i.e., $G^*$) with respect to each individual class $c \in C$.

\textbf{Lemma 3.} \label{lem:3} \textit{The optimization of the global objective $(G \circ F)^*$ with respect to a class $c \in C$, shared by a subset of clients $K'$, is controlled by the optimization of $(G \circ F)_k^*$ for all clients $k \in K'$ on the class $c$.}

\textbf{Proof:} Let $K'$ be the subset of clients in the network sharing class $c \in C$. In our approach, the class-specific weights $(\mathbf{w}_{t,k}^{(G)})_c$ for clients $k \in K'$ are shared with the clients at the end of each communication round, such that:
\begin{align}
    (\mathbf{w}_{t}^{(G)})_c &= mean(\{(\mathbf{w}_{t, k}^{(G)})_c\ \text{if } c \in C_k\ \text{: } \forall k\}) \\
    (\mathbf{w}_{t,k}^{(G)})_c &= (\mathbf{w}_{t}^{(G)})_c\ \forall k
\end{align}
The global sub-objective $G^*$ with respect to the shared class $c$ is controlled by the local optimization of $G_k^*$ on class $c$ for each clients $k \in K'$. Using Lemma \hyperref[lem:2]{2}, we conclude that with respect to shared class $c$, the optimization of global objective $(G \circ F)^*$ is ensured. In other words, regardless of whether all or \emph{some} clients share the class $c$, the $(G \circ F)^*$ is ensured to be optimized with respect to the class $c$.

We further analyze a special case for Lemma \hyperref[lem:3]{3} where $|K'| = 1$, i.e., there is a class $c \in C$ unique to a client $k$.

\textbf{Lemma 4.} \label{lem:4} \textit{The optimization of global objective $(G \circ F)^*$ with respect to a class $c \in C$, unique to client $k$, i.e., it is not shared by any other client in the network, is controlled by the optimization of $(G \circ F)_k^*$ on class $c$. }

\textbf{Proof:} The global sub-objective $G^*$ with respect to the the class $c$ is by definition optimized by the local optimization of $G_k^*$ on class $c$ unique to client $k$. Using Lemma \hyperref[lem:2]{2}, we conclude that with respect to the class $c$, the optimization of the global objective $(G \circ F)^*$ is ensured.

Putting together Lemmas \hyperref[lem:3]{3} and \hyperref[lem:4]{4}, we prove that the optimization of global objective $(G \circ F)^*$ with respect to class $c \in C$ is always ensured and is derived from previously established convergence analyses, regardless of class-heterogeneity in the network \citep{li2019convergence,li2021fedbn}. Therefore, using Lemma \hyperref[lem:1]{1}, we can confer:

\textbf{Theorem 1.} \label{thm:1} \textit{The optimization of global objective $(G \circ F)^*$ is ensured if local objective $(G \circ F)_k^*$ is optimized for all classes $c \in C_k$ using model aggregation strategy $f$ and the optimization is independent of any class-heterogeneity present in the network.}

To conclude, Theorem \hyperref[thm:1]{1} says that regardless of the class-heterogeneity present in the network, the convergence of surgical aggregation is guaranteed. 

\end{document}